\pdfoutput=1
\documentclass[11pt]{article}
\usepackage[preprint]{acl}

\usepackage{times}
\usepackage{latexsym}
\usepackage[T1]{fontenc}
\usepackage[utf8]{inputenc}
\usepackage{microtype}
\usepackage{inconsolata}
\usepackage{graphicx}
\usepackage{fontawesome} 

\title{Benchmarking LLMs in Political Content Text-Annotation: Proof-of-Concept with Toxicity and Incivility Data\thanks{This version was prepared for delivery at the 8th Monash-Warwick-Zurich Text-as-Data Workshop, September 16-17, 2024.}}

\author{Bastián González-Bustamante\thanks{Post-doctoral Researcher in Computational Social Science, Institute of Public Administration, Faculty of Governance and Global Affairs, Leiden University, Netherlands. {\faMapMarker} Wijnhaven, Turfmarkt 99, The Hague 2511 DP, Netherlands. Lecturer, School of Public Administration, Faculty of Administration and Economics, Universidad Diego Portales, Chile. {\faHome} \href{https://bgonzalezbustamante.com/}{https://bgonzalezbustamante.com}, ORCID iD \href{https://orcid.org/0000-0003-1510-6820}{https://orcid.org/0000-0003-1510-6820}.} \\
  Leiden University, Netherlands \\
  Universidad Diego Portales, Chile \\
  \normalsize{\href{mailto:b.a.gonzalez.bustamante@fgga.leidenuniv.nl}{b.a.gonzalez.bustamante@fgga.leidenuniv.nl}} \\
  \\
  \today}

\begin{document}
\maketitle
\begin{abstract}
This article benchmarked the ability of OpenAI’s GPTs and a number of open-source LLMs to perform annotation tasks on political content. We used a novel protest event dataset comprising more than three million digital interactions and created a gold standard that includes ground-truth labels annotated by human coders about toxicity and incivility on social media. We included in our benchmark Google’s Perspective algorithm, which, along with GPTs, was employed throughout their respective APIs while the open-source LLMs were deployed locally. The findings show that Perspective API using a laxer threshold, GPT-4o, and Nous Hermes 2 Mixtral outperform other LLM’s zero-shot classification annotations. In addition, Nous Hermes 2 and Mistral OpenOrca, with a smaller number of parameters, are able to perform the task with high performance, being attractive options that could offer good trade-offs between performance, implementing costs and computing time. Ancillary findings using experiments setting different temperature levels show that although GPTs tend to show not only excellent computing time but also overall good levels of reliability, only open-source LLMs ensure full reproducibility in the annotation.
\end{abstract}

\section{Introduction}

Despite the advantages of digital social media and the Internet, in a broader sense, for collective action and political engagement, an increase in incivility and toxicity in digital interactions has also been observed, becoming a recurrent research topic in recent years \cite[e.g.,][among others]{Schmidt2024, Kim2021, Salgado2023}. In particular, digital interactions and online political discussions seem to be good grounds for incivility and toxicity \citep{Schmidt2024}.

The rapid advancements in AI capabilities since the early 2020s have brought about a paradigm shift in the landscape of computational social science research and the Natural Language Processing (NLP) field. These breakthroughs have revolutionised the traditional text-as-data approach in social sciences, which relied on machine learning using dictionaries, different forms of topic modelling, and supervised, semi-supervised, or unsupervised applications \citep[see also \citealp{GonzalezBustamante2023}]{Watanabe2022}.\footnote{Despite the potential benefits, some disciplines in the social sciences still show a low adoption of automated coding techniques and machine learning approaches. This is primarily due to the pervasive use of resources and time-consuming human-manual data processing, which could be significantly improved by adopting these techniques \citep{Eshima2023, Radford2019}.} By leveraging the zero and few-shot capabilities of Large Language Models (LLMs) and deep learning, we can effectively handle large volumes of content that would otherwise necessitate substantial manual effort or augment and supplement traditional machine learning applications. This novel approach not only provides a distinct avenue to advance our comprehension of toxicity and incivility in digital interactions and political discourse in general but also showcases the potential of LLMs in shaping the future of computational social science research.

That landscape is changing, and at least in massification, an apparent leap forward occurred at the end of 2022 because of the adjusted version of GPT-3 with 175B parameters \citep{Brown2020}. Then, a number of more recent versions, such as GPT-4, in 2023, have shown outstanding performance on different tasks \citep{Cao2024, Gilardi2023, Toernberg2023}. Recently, in September 2024, OpenAI has released their novel o1-preview and o1-mini. Researchers have begun to use OpenAI’s Application Programming Interface (API) for labelling text using various versions of Generative Pre-Trained Transformers (GPTs) for different purposes \citep[][see also \citealp{Gilardi2023, He2024}]{Argyle2023, Gruber2024}.

However, these models may have underlying biases from their training process that may influence the results \citep{WhitePaper2023, Geng2024}. In addition, there are some concerns about dependency on property models owned by for-profit companies, which could jeopardise the reproducibility and openness of research as well as ethical considerations about the transparency and the use of sensitive information without proper consent for the data training process \citep{Spirling2023, Weber2023}. In this context, open-source LLMs have emerged as an alternative to collaborative research, emphasising transparency and reproducibility.

This article involves benchmarking and algorithm auditing a number of models, such as Google’s Perspective algorithm, property or closed-source OpenAI’s GPTs and open-source LLMs, to measure toxicity and incivility on social media during mass protest events. It constitutes a contribution demonstrating the potential of generative artificial intelligence to automate the labelling processes of political content.

\section{Toxicity, Incivility and LLMs for Annotation Tasks}

This section briefly reviews related work on toxicity and incivility and open-source LLMs for annotation tasks. Concerning the former, despite most literature tends to understand the use of disrespectful language as a common feature, there are a variety of conceptualisations that could include vulgarity, profanity, identity-based attacks, and hate speech, among others \citep{Chen2017, Kim2021, Schmidt2024, Schmidt2017, Stoll2023}. 

The text-as-data approach appears as a relevant cornerstone due to the limited number of texts that human coders can annotate, considering time and resources \citep{Schmidt2024}. In this context, the Perspective API, developed by Jigsaw and Google’s Counter Abuse Technology team, is one of the top-shelf options for categorising and classifying toxicity and incivility in the digital sphere. This algorithm was trained on millions of comments from Wikipedia, {\itshape The New York Times} and a variety of other sources labelled by crowdsource raters using distilled Bidirectional Encoder Representations from Transformers (BERT) models into Convolutional Neural Networks (CNNs). Perspective API has been used to moderate media content. For example, \citet{Goyal2022} highlighted the Perspective API as a machine learning solution in the content moderation workflow to tackle online harassment. 

In academic research, to the best of our knowledge, the algorithm has been used to detect toxicity and uncivil comments on Twitter \citep{Hopp2020, Theocharis2020, Orchard2024, Schmidt2024}, Facebook \citep{Hopp2020, Kim2021, Schmidt2024}, news comments \citep{Orchard2024, Schmidt2024} and Wikipedia content \citep{Pavlopoulos2020}.\footnote{It is relevant to bear in mind that there is an important number of works, even including toxicity, that used other transformer models and derivatives from the BERT family, especially the Robustly Optimized BERT Approach (RoBERTa) and distilled or fine-tuned versions, which tend to outperform BERT in several tasks, especially those that involve cross-lingual applications \citep{Timoneda2024}.}

The contribution of this paper is benchmarking Google’s Perspective algorithm, OpenAI’s GPTs and open-source LLMs as classifiers of toxicity, considering a gold standard based on human annotation. Indeed, LLMs are being used in social science in disciplines such as political science to study misinformation, replacing manual processes even instead of traditional NLP approaches and gaining insight into different dimensions of political speech \citep{Linegar2023}. As indicated above, researchers have been using GPTs through OpenAI’s API for a variety of tasks \citep{Gilardi2023, He2024}, and despite different concerns related to reproducibility, privacy and openness of this pay-per-use form, this way tends to offer resources beyond those usually available to the average researcher in social sciences fields, it is straightforwardly to deploy without excessive computational requirements and also GPT-3.5 onwards, especially versions as from GPT-4, tend to excel on a number of tasks and outperform other options \citep{Linegar2023}. 

However, recent studies have shown that open-source LLMs cannot only be used to enhance scientific reproducibility but also may outperform some proprietary models in text annotation and labelling tasks, especially when applying zero-shot classification. Some open-source LLMs are able to perform or even surpass GPT-3.5 and 4 in zero-shot tasks, only lagging in fine-tuned GPTs \citep{Alizadeh2024}. In this sense, using additional information to generate prompt variation, such as context or target data in the pipeline for hate speech detection, could improve the LLMs’ performance by between 20 and 30\% \citep{Roy2023}. Indeed, different prompts and models perform better for different types of content since they can handle better unstructured or structured knowledge \citep[see][]{Zhang2024}.

In addition, some open-source so-called Small Language Models (SLMs), which have seven or fewer billion parameters, thanks to quantisation techniques, are able to perform well and offer a considerable cost reduction compared to larger models or some propriety ones \citep[][see also \citealp{Bucher2024}]{Irugalbandara2024}. Precisely, the capacity to reduce the size of models without sacrificing their performance by fine-tuning and offloading LLMs layers to deploy them locally is a promising ---and sustainable--- avenue of research that opens new possibilities \citep{Hu2021, Linegar2023}.

\section{Data and Methods}

\subsection{Data and Gold Standard}

We used a novel dataset that comprises more than 3.5 million messages posted on Twitter, rebranded as X, about protest events in Argentina ($n = 551,761$) and Chile ($n = 3,125,254$).\footnote{This dataset also contains messages about protests against education budget cuts in Brazil in May 2019 ($n = 1,272,148$), which were excluded from the analysis since our gold standard was created using Spanish native speakers’ human coders.} The Argentinian protests were against coronavirus and judicial reform measures during August 2020, while Chilean messages cover the social outburst stemming from protests against the underground fare hike in October 2019. We scraped all the messages using hashtags in both countries that reached more than 50,000 posts during those months.\footnote{The specific hashtags in Argentina were: \#17AJuntosContraLaImpunidad; \#YoNoMarcho; \#26ATodosAlCongreso; \#1AYoVoy; \#QueSeVayanTodos; \#GobiernoDePayasos. In the Chilean case the hashtags were: \#EstoPasaEnChile; \#ChileDesperto; \#LaMarchaMasGrandeDeChile; \#ChileViolatesHumanRights; \#ChileQuiereCambios; \#EvasionMasivaTodoElDia; \#ChileNoSeRinde; \#EstoNoHaTerminado.} We used the now deprecated academic research access to the Twitter API and downloaded the messages in JavaScript Object Notation (JSON) format.

In order to guarantee the quality of our benchmarking, we created a gold standard using a balanced sample of messages from both countries. First, we classified toxicity in the full corpora using the Perspective algorithm,\footnote{The AUC-ROC score for toxicity is 0.940 for Spanish. The model’s bias was evaluated on synthetic samples of identity groups to test potential underperform. Further information is available on Perspective API documentation for developers at \href{https://developers.perspectiveapi.com}{https://developers.perspectiveapi.com}.} which implied 2,411 hours of computing, and the entire process was deployed on a Raspberry Pi 5, a small computer board with an ARM CPU and meagre power consumption.\footnote{This allowed us to reduce costs and, more importantly, decrease the carbon footprint of our classification task, contributing to greater sustainability and energy efficiency. According to our estimates, our footprint using Raspberry Pi 5 was 6.51 kg of CO2, which implied a reduction of 96\% in emissions that we would have generated running the tasks on an average cloud computing service. It is relevant to note that Perspective API classification included Portuguese messages from Brazil ($n = 1,272,148$) dropped for this benchmarking since the gold standard was elaborated for Spanish, as already mentioned.} Then, we drew a random, balanced sample of 1,000 messages. This sampling process considered the level of toxicity using the Perspective score, which is basically a predicted probability of the algorithm’s classification. Therefore, we balanced the sample by splitting it by country, considering 500 messages from Argentina and 500 from Chile, and then used quintiles of toxicity: $Q_1$ from 0.00 to 0.20; $Q_2$ from 0.21 to 0.40; $Q_3$ from 0.41 to 0.60; $Q_4$ from 0.61 to 0.80; and $Q_5$ from 0.81 to 1.00.

The toxicity was manually annotated by a team of five high-skilled human coders using \citet{Labelbox2024}, an online platform that complies with the most relevant privacy and protection regulations. We used a binary codification performed using the core definitions of Perspective for toxicity as follows: {\itshape “Classify the category of the comment as either TOXIC or NONTOXIC. TOXIC: Rude, disrespectful, or unreasonable comments that are likely to make someone leave the discussion or stop sharing their perspective. NONTOXIC: Civil or nice comments that are unlikely to discourage conversation”}. Each message was hand-coded by two different coders of the team, therefore, we have two ground-truth labels for each sample observation. The inter-coder reliability was excellent: Cohen’s $\kappa$ 0.944 and Krippendorff’s $\alpha$ with a bootstrap of 1,000 iterations 0.944 (95\% CI [0.919, 0.961]). The entire process involved 7.2 hours of annotation and 1.2 hours of revision.\footnote{Further details about the gold standard and the metadata can be found in \citet{GonzalezBustamante2024}. The data are available for replication or reuse.}

\subsection{LLMs for Text-Annotation}

During the gold standard creation process, we applied Perspective API to the whole corpus, therefore, we can compare it with the results of LLMs as classifiers for text annotation. Thus, we ran zero-shot classification tasks to identify toxicity using proprietary, closed-source OpenAI’s GPTs and a number of open-source LLMs. We used the main recent GPTs, namely GPT-4o, GPT-4o mini, GPT-4, GPT-4 Turbo and GPT-3.5 Turbo. We have not included o1-preview and o1-mini, released recently on September 12, 2024, since they are not fully available to all API users at the moment of writing this paper. On the other hand, we locally deployed a selection of open-source LLMs with the minimum temperature level to ensure reproducibility \citep[][see also \citealp{Weber2023}]{Gruber2024}.\footnote{Further information on the computing infrastructure used to deploy locally the LLMs is available in the Appendix.}

Prior to this study, we ran a preliminary benchmark using Perspective API as a proxy of ground-truth labels instead of the gold standard elaborated by human annotators. In these preliminary analyses, available in the Appendix, we used Perspective score with a laxer threshold of 0.55 and a standard of 0.70 and tenfold cross-validated performance and goodness-of-prediction indicators to obtain averaged metrics across the folds for each classifier in order to smooth performance fluctuations. This pre-proof-of-concept (pre-PoC) allowed us to exclude some models that only focused on programming tasks or were explicitly designed for embeddings. We privileged general-purpose models and those that provided a zero-shot classification straightforwardly following our prompt strategy rather than a chain-of-thought, which generally took over one minute per observation. In addition, we prioritised the selection of the state-of-the-art (SOTA) open-source LLMs, though we also included some models prior to Llama 3.1 release that could show good performance and trade-off considering computing time in our pre-PoC results. Therefore, we selected five SOTA open-source LLMs (i.e., Llama 3.1, Hermes 3, Gemma 2 9B, Gemma 2 27B and Mistral NeMo) and five models that could be slightly outdated (i.e., Nous Hermes 2 Mixtral, Nous Hermes 2, Mistral OpenOrca, Orca 2 and Aya) taking into account the accelerated pace of generative AI and NLP.\footnote{Considering the pre-PoC and our final benchmark, we tested 21 open-source LLMs deployed locally. It is important to note that some of the versions of these models with higher number of parameters, such as Llama 3.1 70 and 405B, Hermes 3 70 and 405B, Llama 3 70B, Dolphin Llama 3 70B, among others, are beyond of our infrastructure’s RAM possibilities.}

Our prompt strategy was based on the labelling process by human coders and the core definitions of Perspective and Google. It comprised a query with the same message given to the coders as a system message in order to give context for the classification task. Along with providing texts of our balanced sample, we also listed the categories for the task as follows: {\itshape “Respond with only the category (TOXIC or NONTOXIC). Do not provide any additional analysis or explanation”}.

In sum, we ran a zero-shot classification considering our prompt strategy using five OpenAI’s GPTs (i.e., GPT-4o, GPT-4o mini, GPT-4, GPT-4 Turbo and GPT-3.5 Turbo), five SOTA open-source LLMs (i.e., Llama 3.1, Hermes 3, Gemma 2 9B, Gemma 2 27B and Mistral NeMo) and five ones before the release of Llama 3.1 (i.e., Nous Hermes 2 Mixtral, Nous Hermes 2, Mistral OpenOrca, Orca 2 and Aya) to benchmark all of them against Perspective API and our gold standard based on human annotations.

\subsection{Reproducibility and Temperature Experiments}

We set the temperature at a minimum to ensure our analysis’s reproducibility and tried to avoid LLMs’ hallucinations, as already mentioned. Higher temperatures make LLMs answer more creatively, and although there is no agreement about the role of this parameter on models’ hallucinations, temperature increases are generally tested in hallucination evaluation \citep[see][]{Hao2024}. Thus, we ran some additional classification tasks for the GPTs and open-source models with the best performance in order to test not only reproducibility but also how the randomness of temperature could influence the outcome.

\begin{table*}
  \centering
  \begin{tabular}{lcccc}
    \hline
    \textbf{Model} & \textbf{Accuracy} & \textbf{Precision} & \textbf{Recall} & \textbf{F1-Score }\\
    \hline
Perspective $0.55$ & $0.882$ & $0.975$ & $0.800$ & $0.879$ \\ 
GPT-4o & $0.804$ & $0.735$ & $0.991$ & $0.844$ \\ 
Nous Hermes 2 Mixtral (47B) & $0.829$ & $0.859$ & $0.813$ & $0.835$ \\ 
Aya (35B) & $0.793$ & $0.727$ & $0.979$ & $0.835$ \\ 
GPT-4 & $0.793$ & $0.737$ & $0.953$ & $0.831$ \\ 
Gemma 2 (27B) & $0.785$ & $0.719$ & $0.979$ & $0.830$ \\ 
GPT-4o mini & $0.761$ & $0.695$ & $0.985$ & $0.815$ \\ 
GPT-4 Turbo & $0.757$ & $0.690$ & $0.989$ & $0.813$ \\ 
Nous Hermes 2 (11B) & $0.772$ & $0.727$ & $0.918$ & $0.811$ \\ 
Orca 2 (7B) & $0.773$ & $0.740$ & $0.888$ & $0.807$ \\ 
Mistral OpenOrca (7B) & $0.777$ & $0.790$ & $0.794$ & $0.792$ \\ 
Hermes 3 (8B) & $0.770$ & $0.770$ & $0.811$ & $0.790$ \\ 
Mistral NeMo (12B) & $0.717$ & $0.659$ & $0.976$ & $0.786$ \\ 
Gemma 2 (9B) & $0.697$ & $0.639$ & $0.993$ & $0.778$ \\ 
Llama 3.1 (8B) & $0.706$ & $0.659$ & $0.931$ & $0.772$ \\ 
GPT-3.5 Turbo & $0.667$ & $0.616$ & $0.998$ &  $0.762$ \\ 
Perspective $0.70$ & $0.756$ & $1.000$ & $0.543$ & $0.704$ \\ \hline
  \end{tabular}
  \caption{Goodness-of-Prediction Indicators of Zero-Shot Classifiers for Toxicity Gold Standard}
  \label{tab:performance}
\end{table*}

Consequently, we ran a zero-shot iteration with the temperature at the minimum and then two additional classifications considering low and high-temperature settings at 0.25 and 1.00, respectively, following the test by \citet{Reiss2023}. It is relevant to note that we set the same random number seed for response generation in all our tests. This allowed us to obtain bootstrapped Krippendorff’s $\alpha$ estimates for an iteration under the same settings as the original zero-shot classification and for different temperature levels to test if the lower confidence interval is above the 0.80 threshold considered reliable \citep{Krippendorff2019}. This test also operates as a robustness check of our main analyses.

\section{Results}

\subsection{Benchmark and Error Rate Analysis}

We calculated a variety of standard performance and goodness-of-prediction indicators for Perspective API at a standard of 0.70, the laxer threshold of 0.55 and the zero-shot classifications using the abovementioned GPTs and open-source LLMs. The performance metrics are: (i) accuracy that reports the proportion of correct predictions of the particular classifier in comparison with the human gold standard; (ii) precision that shows the ability of the classifier to identify positive predicted values to identify false negatives; (iii) recall or sensitivity that shows the proportion of correct classifications among true-positive cases; and (iv) F1-score, a combination of precision and recall. Table \ref{tab:performance} presents the indicators per model listed by the F1-score in descending order.

Perspective API, Google’s model distilled from BERT family models, with a laxer cut-off threshold of Perspective score at 0.55, outperforms the GPTs and open-source LLMs we have tested, and it is the closest result to our gold standard with human coders. This finding offers interesting insights since the same model using a standard threshold of 0.70 was ranked at the bottom of the table, presenting one of the lowest performances with an F1-score slightly above 0.70, which is still acceptable. This suggests that the Perspective score as a probability measure tends to be too stringent and should be considered carefully when classifying messages.

Then GPT-4o and Nous Hermes 2 Mixtral, with 47B parameters, show the highest F1-score in classifying toxicity. GPT-4-o was the most advanced, flagship OpenAI’s model available until September 12, 2024, when o1-preview and o1-mini were released, though they are not widely available in the API yet. On the other hand, Nous Hermes 2 Mixtral is an open-source LLM from Nous Research trained on Mixtral and over GPT-4 synthetic data. The F1-score and accuracy of these models are above 80\%. While Nous Hermes 2 Mixtral has a large number of parameters, making it challenging to deploy it on standard computers because of RAM requirements, GPT-4o is deployed directly through the OpenAI’s API, however, as it is a proprietary model could present potential drawbacks already discussed.

\begin{figure*}[t]
  \includegraphics[width=0.96\linewidth]{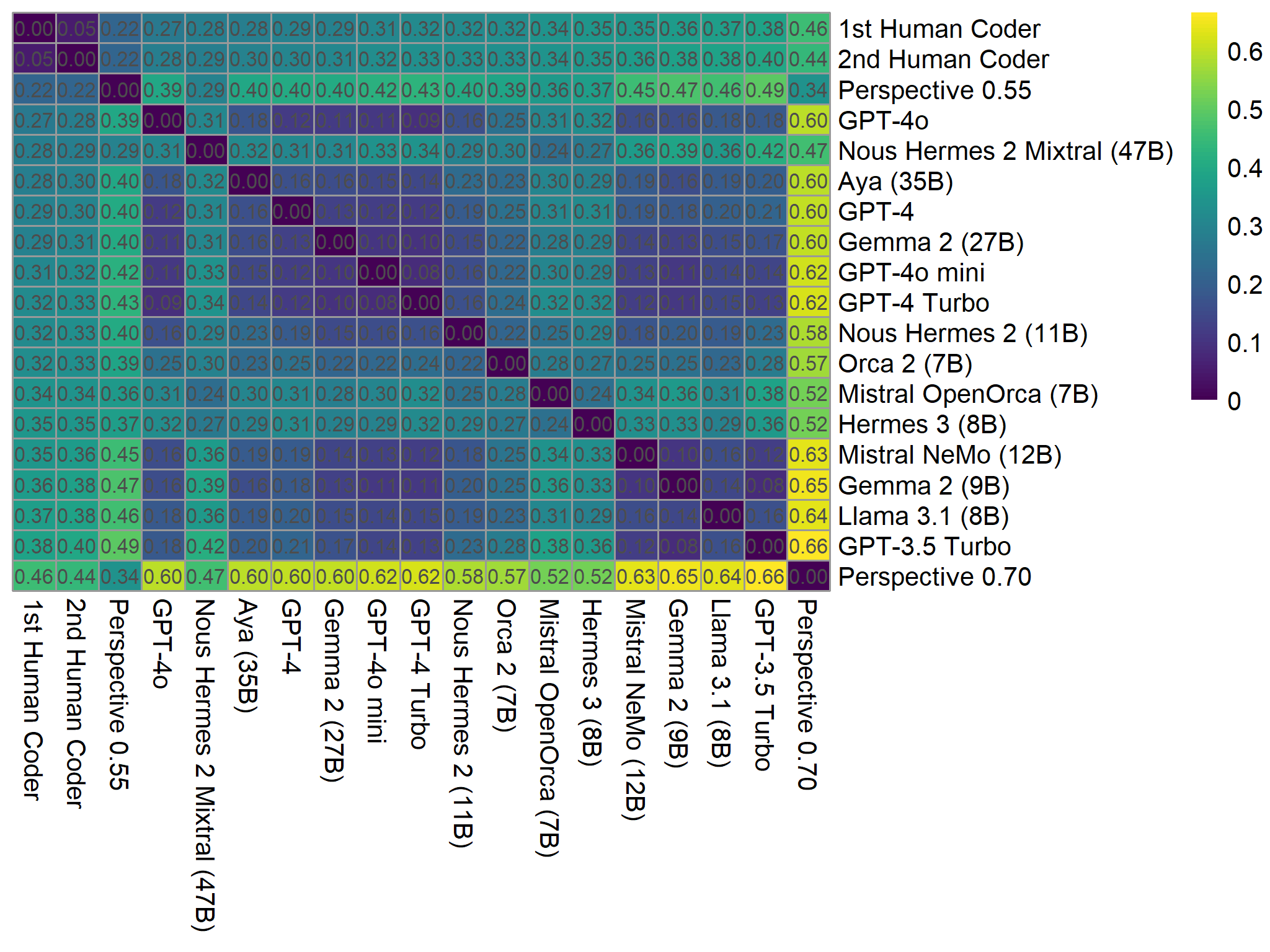}
  \caption {Jaccard Distance Heatmap between Gold Standard, Perspective API and Zero-Shot LLMs Classifiers}
  \label{fig:heatmap}
\end{figure*}

In addition, Aya, with 35B parameters, GPT-4 and Gemma 2, with 27B parameters, also show excellent performance that ranges between 82 and 83\% for F1-score and slightly lower values for accuracy. GPT-4 is the previous generation of OpenAI’s models, while Gemma 2 is a SOTA open-source LLM developed by Google using its own architecture instead of Meta’s, such as most open-source models. On the other hand, Aya is a multilingual open-source LLM that supports 23 languages released before Llama 3.1. 

GPT-4o mini, GPT-4 Turbo, Nous Hermes 2 and Orca 2 also perform in this range. GPT-4o mini, on the one hand, is the most capable small OpenAI’s model, without considering the novel o1-mini, and GPT-4 Turbo is the latest turbo model of its generation. On the other hand, Nous Hermes 2 and Orca 2 could be deployed without significant problems on a standard laptop, and while the former was trained to excel at scientific discussions and coding tasks, the latter is a fine-tuned version of the outdated Meta’s Llama 2 trained by Microsoft to excel in reasoning tasks.

On the contrary, Llama 3.1 and GPT-3.5 Turbo, an outdated GPT for simple tasks, present one of the lowest performances, between 76 and 77\% F1-scores and slightly lower values for accuracy, which are still acceptable metrics. It is important to note that although Llama 3.1 is the novel SOTA model from Meta, this is the lightest version since, as mentioned above, the versions with 70 and 405B parameters are beyond our computing infrastructure.

In Figure \ref{fig:heatmap}, we present a heatmap of Jaccard distances between annotations of the additional human coder, Perspective API classification, GPTs and open-source LLMs. In addition to the high similarity between human coders, two major clusters can be observed. The first group is GPT-4o, Aya, GPT-4, Gemma 2 27B, GPT-4o mini and GPT-4 Turbo, all with high performance. The second cluster is formed by lower-performance models, such as Llama 3.1, Gemma 2 9B, and Mistral NeMo, all SOTA open-source LLMs. 

\begin{figure*}[t]
  \includegraphics[width=0.48\linewidth]{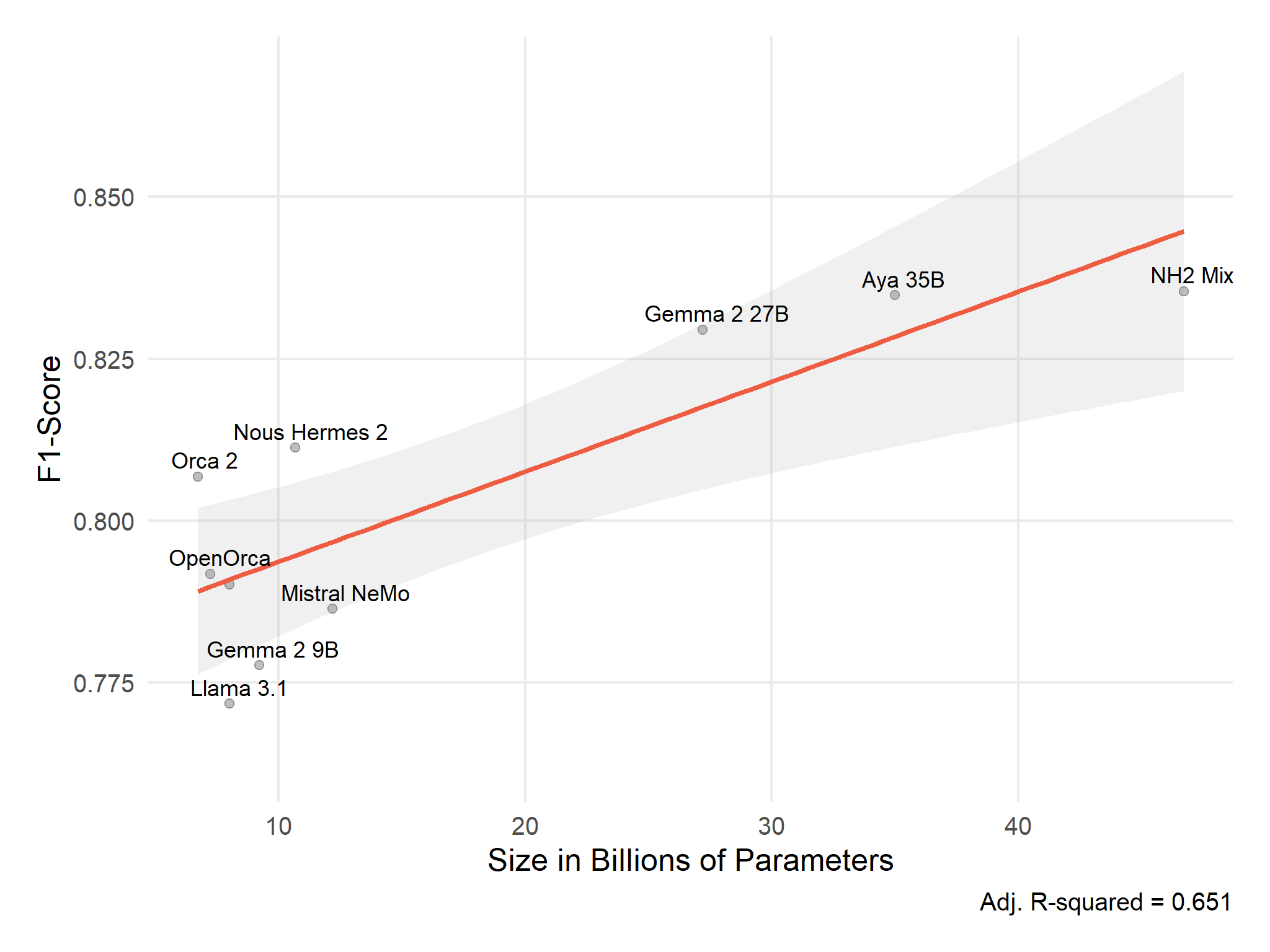} \hfill
  \includegraphics[width=0.48\linewidth]{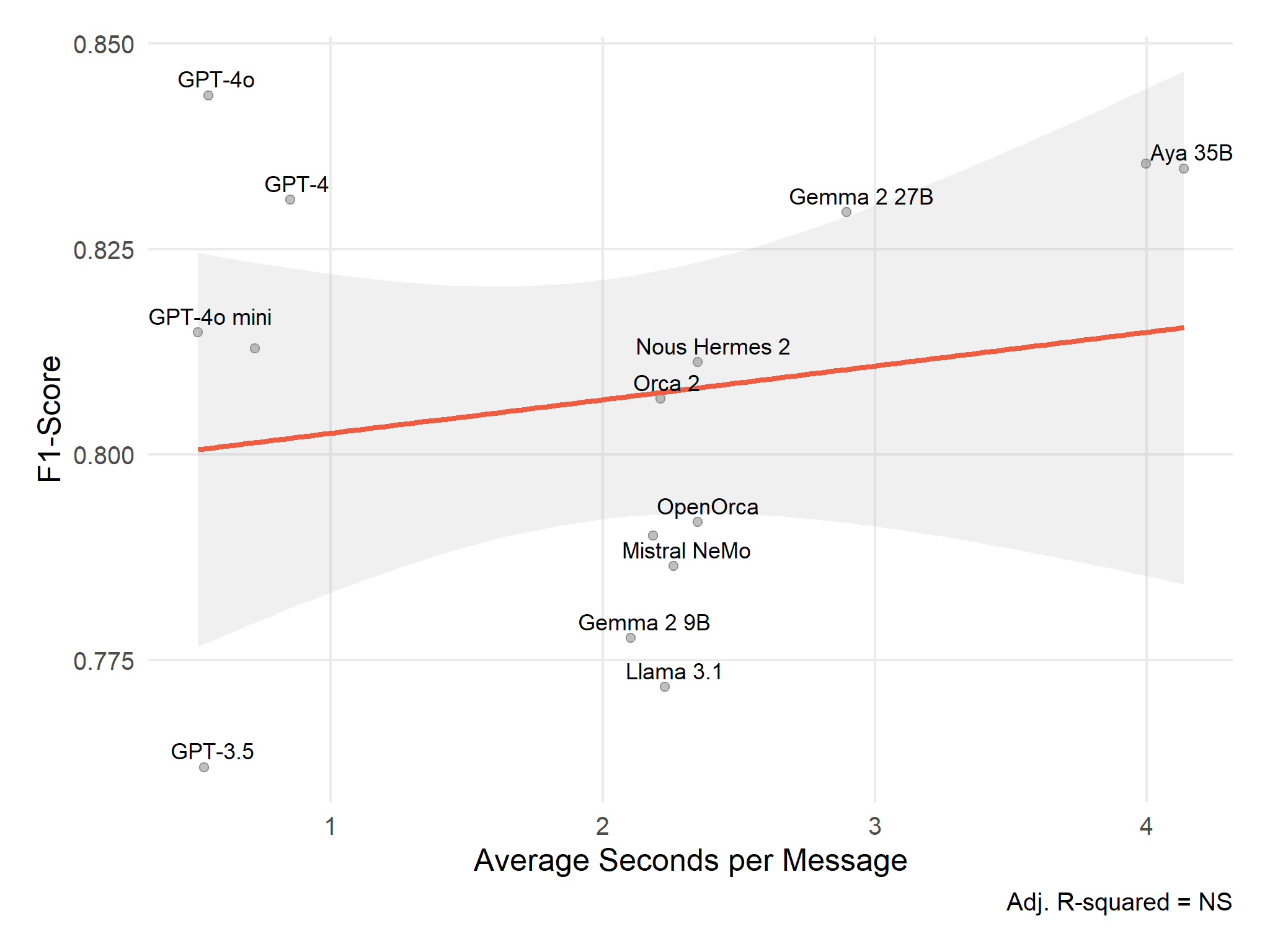}
  \caption {Average Performance, Number of Parameters and Computing Time of Zero-Shot LLMs Classifiers for Toxicity}
  \label{fig:parameters}
\end{figure*}

With an intermediate performance, Mistral OpenOrca and Hermes 3, both with parameters between 7 and 8B, show indexes around 0.30, therefore, they are not highly similar to other models, which could open avenues for ensemble annotations and the use of stacking classifiers. The same applies to Nous Hermes 2 Mixtral, which showed one of the highest performances, as already mentioned. While Mistral OpenOrca is a slightly outdated fine-tuned version of Mistral with 7B parameters trained to outperform models between 8 and 13B parameters, Hermes 3 is the novel SOTA flagship model of the Hermes series of LLMs. 

\subsection{Performance, Parameters and Computing Time}

Figure \ref{fig:parameters} presents the relationship between each classifier’s F1-score, the number of parameters in the case of the open-source LLMs and computing time. While no statistically significant relationship exists between time and performance, the number of billions of parameters is significant ($p = 0.003$). Overall, the models with more parameters tend to show higher performance metrics. 

In this sense, it is possible to identify some interesting cases, such as Nous Hermes 2 and Mistral OpenOrca, which exhibit good performance despite the lower number of parameters. Both also show an average time performance of 2.350 and 2.349 seconds per message, respectively. These values are considerably lower than Aya’s, with 35B parameters and 4.137 seconds per message. It is important to consider that the average human annotation was 15.125 seconds, including labelling and revision time. Perspective API classification, for example, took an average of 1.175 seconds per observation.

In this context, the low average computing time of all GPTs is remarkable: the quickest was GPT-4o mini with only 0.511 seconds per observation, while the slower was GPT-4 with 0.852 seconds, which is still 5x quicker than the average open-source LLMs computing time. In sum, open-source LLMs are almost 6x faster than human coders, while Perspective API and GPTs are even quicker: nearly 13x and 24x, respectively.

\subsection{Reproducibility and Temperature}

\begin{figure*}[t]
  \includegraphics[width=0.9\linewidth]{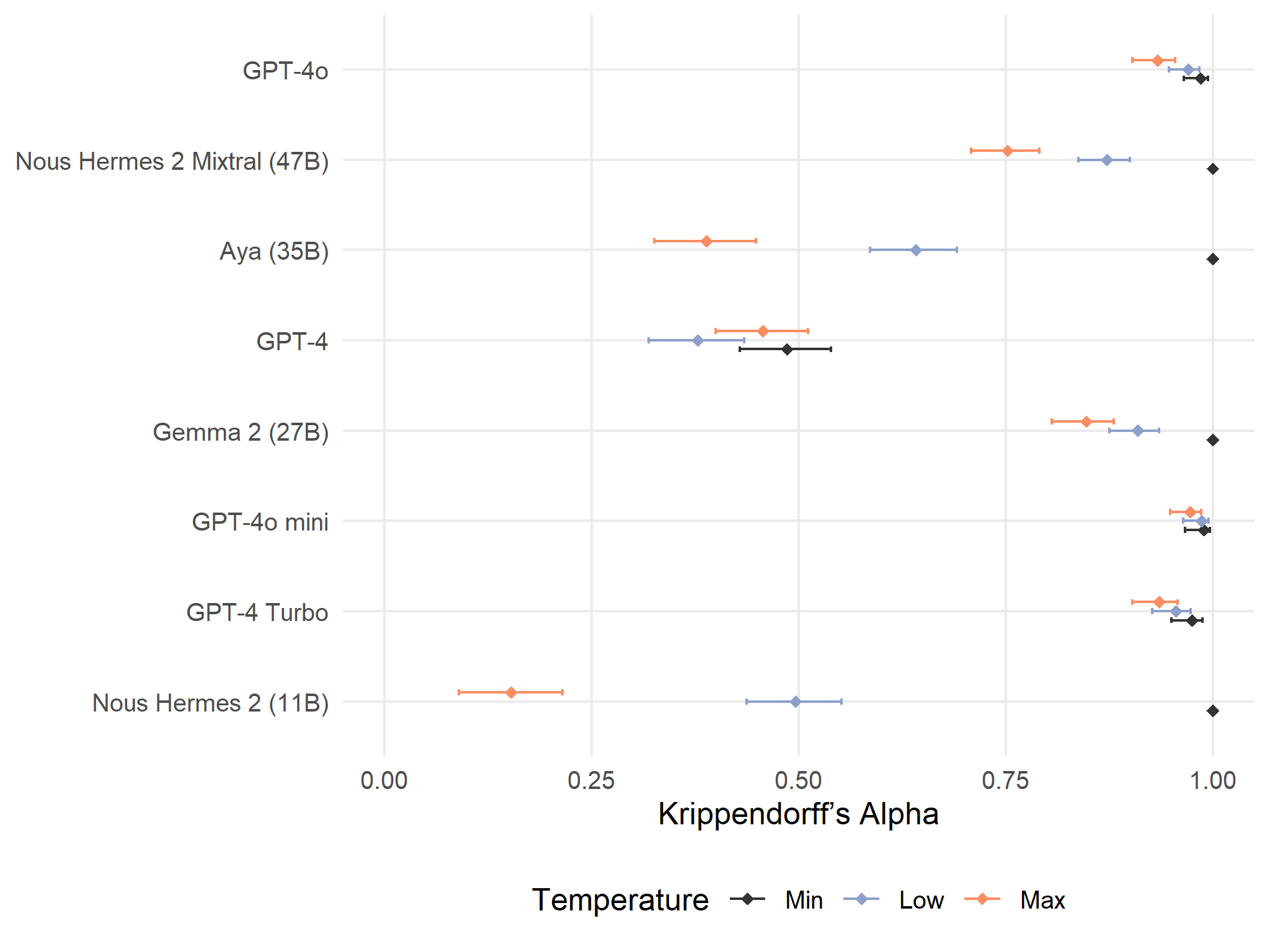}
  \caption {Output Reliability Experiments of Zero-Shot LLMs Classifiers for Toxicity with Best Performance Models}
  \label{fig:experiments}
\end{figure*}

We ran some experiments to test not only the reproducibility but also the influence of different temperature levels on the annotation task, iterating the models with the best performance. Figure \ref{fig:experiments} shows the 1,000 bootstrapped Krippendorff’s $\alpha$ estimates for the original zero-shot classification iteration at the minimum temperature and iterations considering different temperature levels (i.e., 0.25 and 1.00).

While GPTs never ensure entire reproducibility, their reliability is relatively high. Surprisingly, there are no extreme fluctuations setting different temperature levels, except for the case of GPT-4, which suggests that this model should not be used for this type of task. On the other hand, although open-source LLMs tend to show lower reliability under different temperature settings, these models ensure complete reproducibility at the minimum temperature. 

\section{Discussion and Limitations}

This paper offers a benchmarking of Perspective API distilled from BERT family models, some OpenAI’s GPTs and open-source LLMs for annotation and classification tasks on political content, specifically toxicity levels associated with online political discussions during events of protests. Although the models with better performance for this task are Perspective API using a laxer threshold, GPT-4o and Nous Hermes 2 Mixtral with 47B parameters, all the models tested showed F1-scores above 0.70, even some outdated versions such as GPT-3.5 Turbo. It is eye-catching that two of the best options evaluated were deployed locally. Perspective API was executed on a Raspberry Pi 5, allowing a striking carbon footprint reduction of 96\% compared to if we had run it on a standard cloud computing service, and Nous Hermes 2 Mixtral was deployed locally using the Ollama server.

In the case of open-source LLMs, there is a trade-off between the number of parameters, computing time and model performance. Some interesting cases are Nous Hermes 2 and Mistral OpenOrca since both, with a small number of parameters, are able to classify toxicity with good accuracy and faster than larger models. This confirms the possibilities that some small models deployed locally with low costs for classification tasks could offer.

In this context, although GPTs surpass other models in terms of computing time, being 24x faster than human coders and more than 5x more rapid than average open-source LLMs, our experiments setting different levels of temperature showed that only through open-source models deployed locally at a minimum level of temperature is possible to ensure reproducibility in the annotation output.

Although a prominent contribution of this work is that it offers insights into using LLMs for annotation in social science topics, it is relevant to note some limitations. For example, exploring some applications beyond zero-shot classification, such as classification tasks involving few-shot ontologies or chain-of-thought, is possible. In this sense, a potential avenue is incorporating additional context information and prompt variation in the pipeline to improve the classification’s performance, as shown by \citet{Roy2023}.

In sum, this work not only offers insights into the performance of different GPTs and open-source LLMs in annotation tasks of political content and toxicity, specifically offering some guidelines for the application, reproducibility and replicability of these models but also opens new avenues of research on the dynamic’s toxicity and incivility in political phenomena in the digital sphere and for different applications on political speech in broader topics.

\section*{Acknowledgements}


This work was supported by the OpenAI’s Academic Programme and the Faculty of Administration and Economics at the Universidad Diego Portales, Chile. I also thank the Institute of Security and Global Affairs at Leiden University, Netherlands, and the Training Data Lab research group for their support.

\bibliography{Benchmarking_LLMs}

\appendix

\setcounter{table}{0}
\renewcommand{\thetable}{A\arabic{table}}

\section{Appendix}
\label{sec:appendix}

\subsection*{Computing Infrastructure}

The toxicity classification task with the Perspective API was executed locally on a Raspberry Pi 5 with an ARM Cortex-A78 4 Core and 8GB of RAM. Raspberry Pi OS based on Debian GNU/Linux was used.

The open-source LLMs were deployed locally on an Intel Core i9-14900K CPU, NVIDIA GeForce RTX 4070 Super Windforce OC 12GB GPU and 64 GB of RAM (two memories DDR5-4800 of 32GB each). Windows Subsystem for Linux v2.1.5.0 and Ollama v0.1.44 were used for the preliminary analyses, and v0.3.10 for the final ones. It is relevant to note that Ollama suggests having 8GB of RAM to run 7B models, 16GB for 13B and 32GB for the 33B models. The largest models that we ran had almost 47B. An average laptop could run only small and medium models below 7B and between 7 and 13B parameters.

\begin{table*}
  \centering
  \begin{tabular}{lcccc}
    \hline
    \textbf{LLM} & \textbf{Accuracy} & \textbf{Precision} & \textbf{Recall} & \textbf{F1-Score }\\
    \hline
Nous Hermes 2 Mixtral (47B) & $0.841$ & $0.777$ & $0.898$ & $0.831$ \\ 
Nous Hermes 2 (34B) & $0.813$ & $0.767$ & $0.817$ & $0.790$ \\ 
Mistral OpenOrca (7B) & $0.790$ & $0.717$ & $0.870$ & $0.786$ \\ 
Orca 2 (7B) & $0.725$ & $0.624$ & $0.940$ & $0.748$ \\ 
Nous Hermes 2 (11B) & $0.720$ & $0.618$ & $0.947$ & $0.747$ \\ 
Aya (35B) & $0.715$ & $0.608$ & $0.993$ & $0.753$ \\ 
Llama 2 (13B) & $0.691$ & $0.605$ & $0.874$ & $0.712$ \\ 
Notus (7B) & $0.680$ & $0.590$ & $0.889$ & $0.708$ \\ 
Aya (8B) & $0.677$ & $0.577$ & $0.991$ & $0.728$ \\ 
Llama 3 (8B) & $0.640$ & $0.553$ & $0.972$ & $0.703$ \\ 
Qwen 2 (8B) & $0.633$ & $0.545$ & $0.993$ & $0.702$ \\ 
Dolphin Llama 3 (8B) & $0.625$ & $0.542$ & $0.968$ & $0.693$ \\ 
Llama 2 (7B) & $0.579$ & $0.511$ & $0.979$ & $0.671$ \\ 
Gemma (9B) & $0.570$ & $0.506$ & $0.991$ & $0.668$ \\ 
Wizard Vicuna (13B) & $0.472$ & $0.454$ & $1.000$ & $0.623$ \\ 
Qwen (4B) & $0.455$ & $0.446$ & $0.985$ & $0.613$ \\  \hline
  \end{tabular}
  \caption{Goodness-of-Prediction Indicators of Zero-Shot LLMs Classifiers for Toxicity Perspective Score at 0.55 -- Pre-Proof-of-Concept with Perspective API as a Proxy of Ground-Truth}
\end{table*}

\begin{table*}
  \centering
  \begin{tabular}{lcccc}
    \hline
    \textbf{LLM} & \textbf{Accuracy} & \textbf{Precision} & \textbf{Recall} & \textbf{F1-Score }\\
    \hline
Nous Hermes 2 (34B) & $0.762$ & $0.554$ & $0.904$ & $0.683$ \\ 
Nous Hermes 2 Mixtral (47B) & $0.753$ & $0.545$ & $0.956$ & $0.691$ \\ 
Mistral OpenOrca (7B) & $0.711$ & $0.504$ & $0.922$ & $0.650$ \\ 
Orca 2 (7B) & $0.607$ & $0.421$ & $0.962$ & $0.584$ \\ 
Llama 2 (13B) & $0.604$ & $0.416$ & $0.914$ & $0.569$ \\ 
Nous Hermes 2 (11B) & $0.602$ & $0.419$ & $0.974$ & $0.585$ \\ 
Notus (7B) & $0.589$ & $0.407$ & $0.929$ & $0.564$ \\ 
Aya (35B) & $0.570$ & $0.403$ & $1.000$ & $0.574$ \\ 
Aya (8B) & $0.533$ & $0.382$ & $0.996$ & $0.551$ \\ 
Llama 3 (8B) & $0.505$ & $0.368$ & $0.983$ & $0.534$ \\ 
Dolphin Llama 3 (8B) & $0.490$ & $0.361$ & $0.974$ & $0.524$ \\ 
Qwen 2 (8B) & $0.490$ & $0.361$ & $1.000$ & $0.528$ \\ 
Llama 2 (7B) & $0.437$ & $0.338$ & $0.986$ & $0.502$ \\ 
Gemma (9B) & $0.424$ & $0.334$ & $0.990$ & $0.497$ \\ 
Wizard Vicuna (13B) & $0.321$ & $0.299$ & $1.000$ & $0.459$ \\ 
Qwen (4B) & $0.310$ & $0.294$ & $0.983$ & $0.451$ \\ \hline
  \end{tabular}
  \caption{Goodness-of-Prediction Indicators of Zero-Shot LLMs Classifiers for Toxicity Perspective Score at 0.70 -- Pre-Proof-of-Concept with Perspective API as a Proxy of Ground-Truth}
\end{table*}

\end{document}